\documentclass[preprint,5p,times,twocolumn]{elsarticle}

\usepackage[utf8]{inputenc}
\usepackage[T1]{fontenc}
\usepackage{lmodern}
\usepackage{graphicx}
\usepackage{helvet}
\usepackage[figurename=Fig.,labelfont=bf,labelsep=period]{caption}
\usepackage{subcaption}
\usepackage{amsmath}
\usepackage{newtxtext,newtxmath}
\usepackage[colorlinks=true,citecolor=red,linkcolor=black]{hyperref}
\usepackage{algorithm}
\usepackage{algpseudocode}
\usepackage{nicefrac}
\usepackage{tikz}
\usepackage{tabularx}
\usepackage{makecell}
\usetikzlibrary{shapes.geometric, arrows, calc}

\newcommand{\matr}[1]{\mathbf{#1}}

\algnewcommand{\algorithmicand}{\textbf{ and }}
\algnewcommand{\algorithmicor}{\textbf{ or }}
\algnewcommand{\OR}{\algorithmicor}
\algnewcommand{\AND}{\algorithmicand}

\tikzstyle{agent} = [rectangle, rounded corners, minimum width=3cm, minimum height=1cm, text width=3cm, text centered, draw=black]
\tikzstyle{env} = [rectangle, rounded corners, minimum width=3cm, minimum height=1cm, text width=5cm, text centered, draw=black]
\tikzstyle{timesep} = [rectangle, minimum width=1cm, minimum height=1cm, draw=black]
\tikzstyle{arrow} = [thick,->,>=stealth]

\journal{Journal of Water Resources Planning and Management}

\begin{document}
\begin{frontmatter}
\title{Deep Reinforcement Learning for Real-Time Optimization of Pumps in Water Distribution Systems}

\author[HDS]{Gergely Hajgató\corref{cortext}}
\cortext[cortext]{Corresponding author, OrcID: \href{https://orcid.org/0000-0003-4283-126X}{0000-0003-4283-126X}.}
\ead{ghajgato@hds.bme.hu}

\author[HDS]{György Paál}
\author[TMIT]{Bálint Gyires-Tóth}
\date{}

\address[HDS]{Department of Hydrodynamic Systems, Budapest University of Technology and Economics, Budapest, Hungary}
\address[TMIT]{Department of Media Communications and Media Informatics, Budapest University of Technology and Economics, Budapest, Hungary}

\begin{abstract}
Real-time control of pumps can be an infeasible task in water distribution systems (WDSs) as the calculation to find the optimal pump speeds is resource-intensive.
The computational need cannot be lowered even with the capabilities of smart water networks when conventional optimization techniques are used.
Deep reinforcement learning (DRL) is presented here as a controller of pumps in two WDSs.
An agent, based on a dueling deep q-network is trained to maintain the pump speeds, based on instantaneous nodal pressure data.
General optimization techniques (e.g. Nelder-Mead method, differential evolution) serve as baselines.
The total efficiency achieved by the DRL agent compared to the best-performing baseline is above $0.98$, while the speedup is around $2\mathrm{x}$ compared to that.
The main contribution of the presented approach is that the agent can run the pumps in real-time as it depends only on measurement data.
If the WDS is substituted with a hydraulic simulation, the agent still outperforms conventional techniques in search speed.
\end{abstract}

\begin{keyword}
deep reinforcement learning \sep dueling deep Q-network \sep real-time control \sep water distribution systems \sep pump control
\end{keyword}
\end{frontmatter}

\section{Introduction}
Ongoing efforts in the development of online measurement sensors and data acquisition \cite{Nardo2018} let new techniques emerge in real-time control of water distribution systems (WDSs).
The need for smart water networks and economic water supply is motivated mostly by the rapidly growing and urbanizing human population (\cite{urb2018} and \cite{pop2019}).
Ever-growing water networks challenge operators in achieving cost-effective, yet satisfactory operation.
According to the comprehensive literature review of \citet{Mala-Jetmarova2018}, operation optimization is the third most active topic among water supply-related applications.

Operation optimization is often treated as a pump scheduling problem, where dynamic programming (DP) methods are widely utilized because they are able to provide the global optimum if the number of variables is low.
The iterative DP method \cite{Zessler1989} is able to cope with the curse of dimensionality in large WDSs but its application is restricted to convex objective functions.
A more recent work by \citet{Bene2013} introduces a method based on discrete dynamic programming that finds the optimal pump scheduling for medium-sized water networks.
However, the multiple restrictions on the problem definition shows that modern variants of dynamic programming are computationally still too expensive.

\citet{Savic2018} found that many state-of-the-art (SOTA) optimization techniques in WDS operation optimization involves some kind of metaheuristics.
\citet{Ritter2020} presents a multiobjective direct policy search technique where different heuristic approaches share the individuals of a population among different metaheuristics.
This way the different beneficial properties are combined from different metaheuristics but incorrectly chosen hyperprameters of the distinct methods can alter the whole optimization process.

Moreover, these methods require a high amount of hydraulic evaluation which makes the use of surrogate models a reasonable approach.
Surrogate models are approximation models that are computationally cheaper than the corresponding -- physically valid -- numerical simulation.
The input of a surrogate model is usually the same as for the simulation (e.g. boundary and initial conditions) but its output is restricted to just a handful of variables.

Surrogate models with high complexity (e.g. Gaussian process regression, artificial neural networks, k-nearest neighbors) are found to be misleading for non-convex objective functions as these smooth the response surface that leads to suboptimal solutions \cite{Simpson2004}.
In concordance, \citet{Razavi2012} shows for water network simulations that substitutes with less complexity are more robust in cases where an unseen part of the explanatory variable space has to be discovered.
\citet{Machell2010} introduces a hybrid approach where, the governing equations of the hydraulic simulation are supported with an extended number of boundary conditions that reduces the computation cost of the hydraulic model.
The additional information is extracted from online sensory data including pressure and flow rate measures.

Optimization convergence can be enhanced by improving the optimum search itself to reduce the time of the overall process.
A promising branch of these efforts is about finding the optimal parameters of the optimization algorithm.
These parameters are called hyperparameters, and methods that optimize these hyperparameters during the single optimization process are referred to as hyperheuristic methods \cite{Savic2018}.
\citet{MccLymont2013} present a Markov chain-based approximator that is superior in finding optimal hyperheuristics compared to Simple Random and TSRoulWheel methods.

Though heuristics are robust in finding the global optima, the computational burden is still too high for real-life WDSs.
Model predictive control (\cite{Pascual2011} and \cite{Wang2017}) overcomes the computational issue by using a two-layer optimization approach where the layers predict optimal ON-OFF schedules for pumps and valves in different time horizons.
The method is able to run large WDSs in real-time, the capabilities are presented on the example of the water network of Barcelona.
Iterative extended lexicographic programming (iELGP), introduced by \citet{Abdallah2017} and \citet{Abdallah2019} copes even with continuous decision variables like the set-point of variable speed pumps (VSPs) and chlorine dosing machines.
With iELGP, \citet{Abdallah2019} achieved SOTA performance on the C-Town benchmark problem wherein both fixed-speed and variable speed pumps were presented.

Water distribution system operation also benefits from recent advances in machine learning.
Artificial neural networks are used in a wide variety of applications from demand forecast \cite{Guo2018} through attack detection \cite{Taormina2018} to pump scheduling \cite{Bhattacharya2003}.

\citet{Bhattacharya2003} uses an ANN-based reinforcement learning agent as an optimal controller for switching on or off pumps in a water network.
The input of the ANN is water levels in reservoirs besides the water demand in the WDS, while its output is the decision about a pump to operate it or not.
The neural network is trained with reinforcement learning (RL) in a manner the RL agent is encouraged to replicate the moves of the master controller algorithm.

\citet{Mahootchi2007} introduces an agent, based on reinforcement learning for the control of one reservoir in a water distribution network and finds that the RL approach achieves comparable results to stochastic dynamic programming in the benchmark case.

A sophisticated optimization technique is introduced by \citet{Castelletti2013} for multiobjective tasks, where the importance of the standalone objectives is unknown or uncertain at the beginning of the optimization.
Recombining these objectives yields a different optimal solution, hence a movement on the Pareto frontier.
\citet{Castelletti2013} approximate the optimal policy with fitted q-iteration (FQI) during an optimization run but also add the weights of the standalone objectives to the state space of the environment used in FQI.
The approximation of the optimal policy is constructed as the function of the importance weights, and the frontier can be drawn from one single run with sampling the weight space as well.

All the pump scheduling methods above rely on the simulation of the water distribution system that limits their usability when the size or the complexity of the water network is too large.
The most resource-intensive method is finding optimal pump speeds for a time step since the descriptive variables are continuous and the evaluation of every trial needs to run the hydraulic simulation as well.

A reinforcement learning agent for real-time control for variable speed pumps is presented in this paper.
The pump scheduling problem is omitted and only the optimum speed search for a given demand-distribution is realized in a smart water network.
A significant novelty of the method is that the hydraulic simulation of the WDS is not needed for the controller because it relies only on measurement data.
Moreover, the training is carried out in such a way that the agent is able to develop different policies compared to the reference, so that the agent can outperform the master algorithm it was trained with.
If online pressure data are not available, the hydraulic simulation of the water distribution system can be used as a surrogate of the real WDS, like in the case of existing optimization methods.
In such cases, the agent-based control is able to outperform current optimization techniques in the number of hydraulic model evaluation.
As the proposed technique is able to cope with multiple objectives with a simple weighted sum of objectives, it can serve as the optimal pump speed search algorithm in hybrid pump schedule optimization solutions.

The paper is organized as follows. First, the simplified problem is introduced including the reference methods. The details of reinforcement learning methodology are described thereafter. Then, the settings of the case study, including the water distribution networks and the training settings are presented. The performance of the agent is evaluated on the WDSs and the conclusions of the training are drawn. Finally, the key findings and possible applications are mentioned.

\section{Methodology for pump speed control with reinforcement learning}
\subsection{Problem definition and reference solutions}
The present study aims at providing optimal set-points for variable speed pumps in a WDS according to given nodal pressures.
Hence, the conventional pump scheduling problem is simplified and the problem is formulated as follows.

Instantaneous operation point optimization is carried out neglecting temporal changes in the demand.
The competing algorithms have to find the optimal pump speeds in terms of hydraulic efficiency with respect to randomly generated demands.
Besides, the inflow and outflow into and from the tanks should be minimized, and pressure (head) should be kept in a specific range in all of the nodes.
Customer satisfaction implies a lower pressure limit, while the mechanical strength of water fittings and other elements of the water network implies an upper limit as well.

These objectives are formulated into a simple weighted sum of objectives.
The meaning of this combined value is two-fold, as it serves as an objective value for the wide-spread optimization algorithms and is interpreted as the value of the state in the Markovian decision process described later.
The rating of the final state is the same regardless, of whether it is achieved by the reference optimization methods or the reinforcement learning-based agent.

The calculation depends on the observation of the environment that is the ensemble of the state variables and some other quantities that are not involved in the decision-making but are measurable.
The algorithm of the calculation is presented in Alg.~\ref{alg:state-value}.
The input of the algorithm are the state variables $\matr{S}$, the set of the pump efficiencies $\matr{E}_\mathrm{pump}$, the volumetric flow rate $\matr{F}_\mathrm{pump}$ delivered by the pumps and the signed volumetric flow rate $\matr{F}_\mathrm{tank}$ of the tanks.

\begin{algorithm}
\caption{Calculating the state value $v(\matr{S})$}
\label{alg:state-value}
\begin{algorithmic}[1]
    \Function{CompStateValue}{$S, E_\mathrm{pump}, F_\mathrm{pump}, F_\mathrm{tank}$}
        \State $\matr{H}=\{h_{1},\dots,h_{N}\} \gets \Call{ExtractNodalHeads}{\matr{S}}$
        \State $n_\mathrm{wrong} \gets 0$
        \ForAll {$h \in \matr{H}$}
            \If{$h < h_\mathrm{min} \OR h > h_\mathrm{max}$}
                \State $n_\mathrm{wrong} \gets n_\mathrm{wrong}+1$
            \EndIf
        \EndFor
        \State $c_\mathrm{tot} \gets \sum{\matr{F}_\mathrm{pump}}+\sum{\matr{F}_\mathrm{tank}}$
        \State $\eta_\mathrm{tot} \gets \prod{\matr{E}_\mathrm{pump}}$
        \State $r_\mathrm{feed} \gets \frac{c_\mathrm{tot}}{c_\mathrm{tot}+\sum_{i=1}^{{\#}\matr{F}_\mathrm{tank}}{\vert{f_i}\vert}}$
        \State $r_\mathrm{satisfaction} \gets 1-\frac{n_\mathrm{wrong}}{\#\matr{H}}$
        \State $r_\mathrm{eff} \gets \frac{\eta_\mathrm{tot}}{\eta_\mathrm{limit}}$
        \State $v \gets w_\mathrm{satisfaction} \cdot r_\mathrm{satisfaction} + w_\mathrm{eff} \cdot r_\mathrm{eff} + w_\mathrm{feed} \cdot r_\mathrm{feed}$
        \State \Return $v$
    \EndFunction
\end{algorithmic}
\end{algorithm}

The state value $v$ is a weighted combination of $3$ objectives.
\begin{enumerate}
     \item $r_\mathrm{satisfaction}$ represents the satisfaction of the consumers that is characterized by the number of the insufficiently supplied nodes $n_\mathrm{wrong}$ wherein the head is out of the acceptable range bounded by $h_\mathrm{min}$ and $h_\mathrm{max}$. The number of problematic nodes is divided by the number of all nodes in the water distribution system, and the ratio is subtracted from $1$. The outcome is the basis of the first weighted objective, the satisfaction ratio.
     \item $r_\mathrm{eff}$ represents the efficiency of the pumps, and it is just the product of the standalone pumps feeding the network divided by the product of the theoretical peak efficiencies of all of the pumps.
     \item $r_\mathrm{feed}$ represents the feed ratio of the water network that is solved by the pumps supplying water from the wells or other sources, on the one hand and by the tanks and reservoirs, on the other hand.
\end{enumerate}
In the current experiment, using the water reserve is considered expensive, and using the reserve is punished by a lower objective value, which is a simplification of the problem.
However, this methodology is easily changeable by adding the target inflows and outflows to the state value and comparing the actual ones to those.
In this way, the agent would be forced to learn to maintain the water flux of the tanks but this is a harder problem to solve.
For the time being, this is ignored by forcing the agent to minimize the water flux of the tanks.
Thus, the feed ratio is the amount of total consumption divided by the sum of the total consumption and the water flux of the tanks.

The standalone objectives are also presented in Eq.~\ref{eq:objectives}, where the notations are the same as for Alg.~\ref{alg:state-value} except for $f_{\mathrm{tank}}$ that denotes the sum of the absolute value of inflow/outflow for every tank.

\begin{equation}\label{eq:objectives}
    \begin{split}
        r_{\mathrm{satisfaction}} &= 1 - \frac{n_{\mathrm{wrong}}}{n_{\mathrm{nodes}}},\\
        r_{\mathrm{eff}} &= \frac{\eta_{\mathrm{tot}}}{\eta_{\mathrm{limit}}},\\
        r_{\mathrm{feed}} &= \frac{c_{\mathrm{tot}}}{c_{\mathrm{tot}}+f_{\mathrm{tot, tank}}}.
    \end{split}
\end{equation}

The above formulation of the standalone objectives leads to $3$ measures, whose maximum values are $1.0$ all.
These are then weighted separately by constant factors.
The coefficients in the presented experiments are chosen arbitrarily $\nicefrac{8}{16}$, $\nicefrac{5}{16}$ and $\nicefrac{3}{16}$ as the ratio of the satisfaction, the efficiency and the feed, respectively.

The task was solved with different optimization techniques to provide reference: fixed step-size random search (FSSRS), particle swarm optimization (PSO), differential evolution (DE), and Nelder-Mead method.
PSO, DE and Nelder-Mead method are able to solve operation point optimization of VSPs, see the work of \citet{Baltar2008}, \citet{Odan2015} and \citet{Celi2017}, respectively.

A suboptimal optimization method was also applied on the task that is referred to as one-shot random trial (One-shot RT).
This method sets pump speeds randomly inside the feasibility domain and it serves as a baseline: the reinforcement learning agent has to be able to outperform this approach if it can interpret any information on the connection of nodal pressures and optimal pump speed settings.

\subsection{Dueling deep Q-network agent as a real-time controller for water networks}
The drawback of the reference methods is the incapability to accomplish real-time control because of the necessity of computationally expensive simulation of the hydraulics.
In this paper, a dueling deep q-network agent is introduced to optimize the speeds of the pumps in a water distribution system.
The agent relies on simulation data in the training phase only; thus, the controlling is feasible real-time, regardless of the size of the water network and the number of pumps controlled.

The proposed technique is based on Markov decision processes (MDPs) that are key building blocks of reinforcement learning.
In the terminology of MDPs an \textit{agent} works in an \textit{environment} by taking \textit{actions} according to a  \textit{policy}.
The environment is observable to the agent through its \textit{state} that changes according to the actions.
The instantaneous quality of a state is measured by the \textit{reward} that is signaled to the agent.
Rewards of a decision sequence are combined in the \textit{value function}.
The value function measures the overall quality of the starting state in the decision sequence according to the given policy.
The goal of the learning is to find a policy with which the agent can govern the environment to the most advantageous state from any initial state.

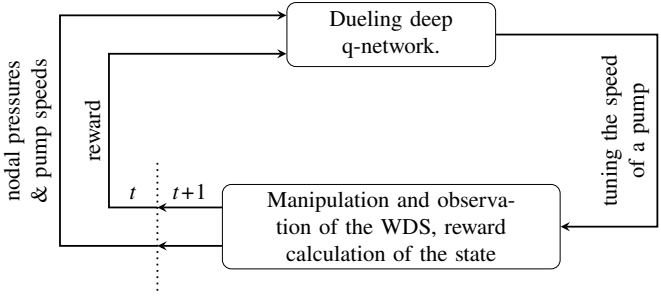
\begin{figure}
    \centering
    \def\nodeSep{3cm}
    \def\timeSep{1cm}
    \def\rewSep{.3cm}
    \resizebox{\linewidth}{!}{
    \begin{tikzpicture}
        \node (agent) [agent] {
            {Dueling deep q-network.}
            };
        \node (env) [env, yshift=-{\nodeSep}] {
            {Manipulation and observation of the WDS, reward calculation of the state}
            };
        \draw [thick, dotted] ($(env.west)+(-\timeSep,-1cm)$) -- ($(env.west)+(-\timeSep,1cm)$);
        \draw [arrow] (agent.east)
            -- ($(env.east)+(1.5cm,\nodeSep)$)
            -- node[anchor=south, rotate=90, align=center, text width=\nodeSep] {tuning the speed of a pump} ++(0,-{\nodeSep})
            -- (env.east);
        \draw [arrow] ($(env.west)+(0,+\rewSep)$)
            -- node[anchor=south, align=center] {$t\!+\!1$} ++($(-\timeSep,0)$);
        \draw [arrow] ($(env.west)+(0,-\rewSep)$)
            -- ++($(-\timeSep,0)$);
        \draw [arrow] ($(env.west)+(-\timeSep,+\rewSep)$)
            -- node[anchor=south, align=center] {$t$} ++(-.75\timeSep,0)
            -- node[anchor=south, rotate=90, align=center] {reward} ++($(0,\nodeSep-2*\rewSep)$)
            -- ($(agent.west)+(0,-\rewSep)$);
        \draw [arrow] ($(env.west)+(-\timeSep,-\rewSep)$)
            -- ++(-1.5\timeSep,0)
            -- node[anchor=south, rotate=90, text width=\nodeSep, align=center] {nodal pressures \& pump speeds} ++($(0,\nodeSep+2*\rewSep)$)
            -- ($(agent.west)+(0,+\rewSep)$);
    \end{tikzpicture}
    }
    \caption{The agent-environment ecosystem tailored to the controlling of the pumps in a  water distribution system.}
    \label{fig:rl-wds}
\end{figure}

In the present study, the agent is a dueling deep q-network and it interacts with the environment that is based on the water network.
The decision-making sequence and most of its concepts are depicted in Fig.~\ref{fig:rl-wds} (originating from \citet{Sutton2018}) where $t$ and $t+1$ refer to two consecutive steps (also referred to as time steps) in the sequence.

According to Fig.~\ref{fig:rl-wds}, the communication between the environment and the agent during the training is as follows.
\begin{enumerate}
    \item The environment provides its state variables -- the ensemble of the nodal pressures and the pump speed ratios -- to the agent in a concatenated vector. Speed ratios are the actual pump speed divided by the nominal pump speed.
    \item The agent maps the state values to an action according to its current policy and executes it in the environment. The action can be the increase or decrease of one of the pump speed ratios or keeping the speed ratios as they were set in the previous step. The latter is referred to as \textit{siesta}.
    \item The new hydraulic properties are provided by the hydraulic simulator, and the value of the new state is calculated. Besides, the reward is calculated from the state value and some other properties of the environment.
    \item The environment propagates the new state and the corresponding reward to the agent to induce the next move.
    \item At last, (1) the preceding state, (2) the preceding action taken, (3) the corresponding reward, and (4) the new state is stored as a training sample wherewith the action-value estimator is updated later.
\end{enumerate}

This cyclic process goes on until the environment reaches a terminal state, which can be caused by two distinct events.
On the one hand, the agent is allowed to run until a fixed number of steps to get to the most advantageous state.
The environment keeps track of the number of steps taken, and it terminates the episode when the agent reaches the limit.
On the other hand, the agent can decide to do nothing instead of changing one of the speed ratios.
If the agent adheres to a state for a predefined number of consecutive steps, the episode is terminated as well.

Seeking to achieve the optimal policy leads to the grading of different policies.
A policy is regarded better than another one if the value function is equal or higher for all states by taking actions under that one.
The value function can be formulated to depend on different variables, the related form is the action-value function, that is for policy $\pi$:
\begin{equation}\label{eq:action-value_fun}
    q_{\pi}(s, a) \doteq \mathbb{E}_{\pi} \left[ \sum_{k=t+1}^{T}{\gamma^{k-t-1} R_{k}} \biggr\rvert S_{t}=s, A_{t}=a \right] \;.
\end{equation}

The value is represented as an expected value of the sum of rewards $R_k$ in the forthcoming sequence started from state $S_t$ and taking action $A_t$ under policy $\pi$.
The importance of future rewards can be attenuated with the discount-factor $\gamma$, such that $\gamma \in (0, 1]$.

If the agent runs under the optimal policy, its actions maximize the reward in every state-transition; thus, the optimum of the action-value function is:
\begin{equation}\label{eq:max_action-value}
    q_{\pi^{*}}(s, a) \doteq \mathbb{E}_{\pi^{*}} \left[ R+\gamma\max_{a'}q^{*}(s', a') \biggr\rvert S_{t}=s, A_{t}=a \right] \;,
\end{equation}
where the forthcoming action and state is denoted by a prime and $q^{*}$ refers to action-values under the optimal policy.

Eq.~\ref{eq:max_action-value} can be solved in different ways, for the relevant techniques see \citet{Sutton2018}.
SOTA solutions use artificial neural networks as value function estimators to cope with the complexity of the equation.
ANNs are good candidates for the task, because they are universal function approximators as shown by \citet{Cybenko1989}.

A variant of deep q-networks (DQN, introduced by \citet{Mnih2013}) is chosen in the present study to approximate the action-value function.
The term deep refers here to the fact that the neural network architecture has two or more hidden layers.
The agent can be trained by minimizing the loss
\begin{equation}\label{eq:dqn}
    L_{i}(\theta_{i}) = \mathbb{E}_{\pi} \left[ (r+\gamma\max_{a'}q(s',a',\theta_{i-1}) - q(s,a,\theta_{i}))^2 \right]\;,
\end{equation}
where $\theta_{i}$ corresponds to the weights in the ${i}^{th}$ update of the deep q-network (a deep neural network in this case) and $L_{i}$ is the mean squared error defined between the $(i-1)^{th}$ and $i^{th}$ approximation of the value function.
A significant advantage of Eq.~\ref{eq:dqn} over Eq.~\ref{eq:max_action-value} is that it does not have to evaluate a whole decision sequence to get a new assumption on the value of the updated state; hence, it can be trained off-policy.

Eq.~\ref{eq:dqn} is optimized with stochastic gradient descent method as it is a general, yet effective optimization algorithm for training ANNs \cite{bottou-bousquet-2008}.
Moreover, the last layer of the deep q-network is forked: one layer approximates the value of the current state, while the other layer estimates the advantage in the value by taking the action.
This is intended to speed up training convergence and the technique is called dueling deep q-networks, as introduced by \citet{Wang2015}.

\begin{algorithm}[ht]
\caption{Step procedure of the agent}
\label{alg:step}
\begin{algorithmic}[1]
    \Procedure{Step}{$a$}
        \State $n_\mathrm{siesta} \gets 0$
        \State $n_\mathrm{steps} \gets n_\mathrm{steps}+1$
        \State $terminal \gets n_\mathrm{steps} = N$
        \State{$r \gets penalty$}
        \If{$a\not=siesta$}
            \State{$\matr{PS} \gets \Call{UpdatePumpSpeeds}{a}$}
            \State{$\delta \gets \lVert{\matr{PS}-\matr{PS}_\mathrm{ref}}\rVert$}
            \If{$any(\matr{PS}<ps_\mathrm{lo}) \OR any(\matr{PS}>ps_\mathrm{hi})$}
                \State{$\matr{PS} \gets \Call{RevertAction}{a}$}
            \Else
                \If{$\delta<\delta^*$}
                    \State{$r \gets \delta \cdot constant$}
                \EndIf
                \State{$\matr{S} \gets \Call{UpdateState}{a}$}
                \State{$\matr{E}_\mathrm{pump}, \matr{F}_\mathrm{pump}, \matr{F}_\mathrm{tank} \gets \Call{Observe}{\matr{S}}$}
                \State{$v \gets \Call{CompStateValue}{\matr{S}, \matr{E}_\mathrm{pump}, \matr{F}_\mathrm{pump}, \matr{F}_\mathrm{tank}}$}
                \State{$\delta^* \gets \delta$}
            \EndIf
        \Else
            \State{$n_\mathrm{siesta} \gets n_\mathrm{siesta}+1$}
            \If{$n_\mathrm{siesta}<3$}
                \If{$1-{v}/{v_\mathrm{ref}}<\epsilon$}
                    \State{$r \gets n_\mathrm{siesta} \cdot constant$}
                \EndIf
            \Else
                \If{$1-{v}/{v_\mathrm{ref}}<\epsilon$}
                    \State{$r \gets bonus$}
                \EndIf
                \State{$terminal \gets \texttt{True}$}
            \EndIf
        \EndIf
        \State \Return $r, \matr{S}, terminal$
    \EndProcedure
\end{algorithmic}
\end{algorithm}

A crucial component of the reinforcement learning setting is the rewarding of the agent as it influences the behavior of the trained agent.
Alg.~\ref{alg:step} summarizes the step procedure with the rewarding system of the present study.

The reward is based upon the state value but it is not a direct match, as other goals have to be reached beside optimality, as shown in Alg.~\ref{alg:state-value}.
As the control is realized stepwise, the agent has to be taught to set the optimal speeds in as few steps as possible.
The basic idea behind the presented step function, including the reward calculation is as follows.
At the beginning of every new scenario, i.e., after the generation of a new demand-distribution, the optimal setting of the pump speeds is obtained by the Nelder-Mead method.
This speed setting $\matr{PS}_\mathrm{ref}$ and the corresponding state-value $v_\mathrm{ref}$ is considered as a reference, guiding the agent softly in the search space.
After the agent intervenes, the reward is based upon the closeness $\delta$ of the reference solution and the agent's solution in the search space, defined by the feasible pump speed ratios.
At this stage, the agent gets a reward $r$ if it gets closer to the reference solution on a monotonous trajectory, i.e., it does not make steps increasing the distance $\delta$ between the solutions.
The agent can also choose to make a \textit{siesta}; in this case, the reward is computed differently.
In siestas, the actual value of the state is computed according to Alg.~\ref{alg:state-value} and compared to the reference value.
If the current state value is higher than the $98\%$ percent of the reference value (i.e., the value of $1-v/v_\mathrm{ref}$ is at most $\epsilon$), the agent is rewarded with a slightly higher score.
The immediate reward achieved in the first siesta is doubled if the agent chooses to stay in this beneficial state for the second time.
Finally, the episode is terminated after the third consecutive siesta, and the agent gets an extra reward that is significantly higher than the rewards that are achievable with the active moves.
In contrast, the agent gets a negative reward whenever it recedes from the reference speed setting with an active action or if it stays in a suboptimal position, i.e., if it makes a siesta in a state, with a value below $98\%$ of the reference.
A penalty is given even when the agent tries to set pump speeds lower or higher as the feasibility limits $ps_\mathrm{lo}$ and $ps_\mathrm{hi}$, respectively.
With this concept, the agent is incited to move along the optimal position continuously and terminate the episode there.
The $2\%$ tolerance in the comparison of the state values is needed because of the different variable space from where the solution is acquired with the reference method and with the reinforcement learning agent: the Nelder-Mead method chooses candidate solutions from a continuous variable space while the dueling deep q-network handles a discrete action space; thus only a restricted set of pump speeds is available compared to Nelder-Mead.
Although the discretization can be arbitrarily fine, the dimensionality of the q-function -- hence, the length of the training of its estimator -- depends on the size of the discrete parameter space.

The soft guidance means in the above process that the agent can achieve high rewards even far from the reference solution if it has a similar value.
This is encouraged during training by taking random actions in a part of the whole process, thanks to the $\varepsilon$-greedy policy.

After training, the agent relies only on the observation of the environment to take its next action; thus, the calculation of the reward -- including the computationally expensive reference calculation -- is completely omitted.
 
\section{Case study}
\subsection{Reference water distribution systems}
The proposed controller is trained and tested on the modified Anytown and D-Town (referred to as C-Town in some literature) water distribution systems that are common among the WDS design and management research community.
These networks serve as benchmarks for water network design methods wherewith the networks can be modified and extended to comply with increased water needs and expanding consumption areas.
The original topology of both WDSs is accessible online from the web page of the Centre of Water Systems research group at the \citet{cwsExeter}.
Moreover, the numerical models of the WDSs are also available in an \verb+EPANET+-compatible format.
The modified versions are referred to as Anytown-mod and D-Town-mod in the following.

\begin{figure}[ht]
    \centering
    \includegraphics[width=\linewidth,keepaspectratio]{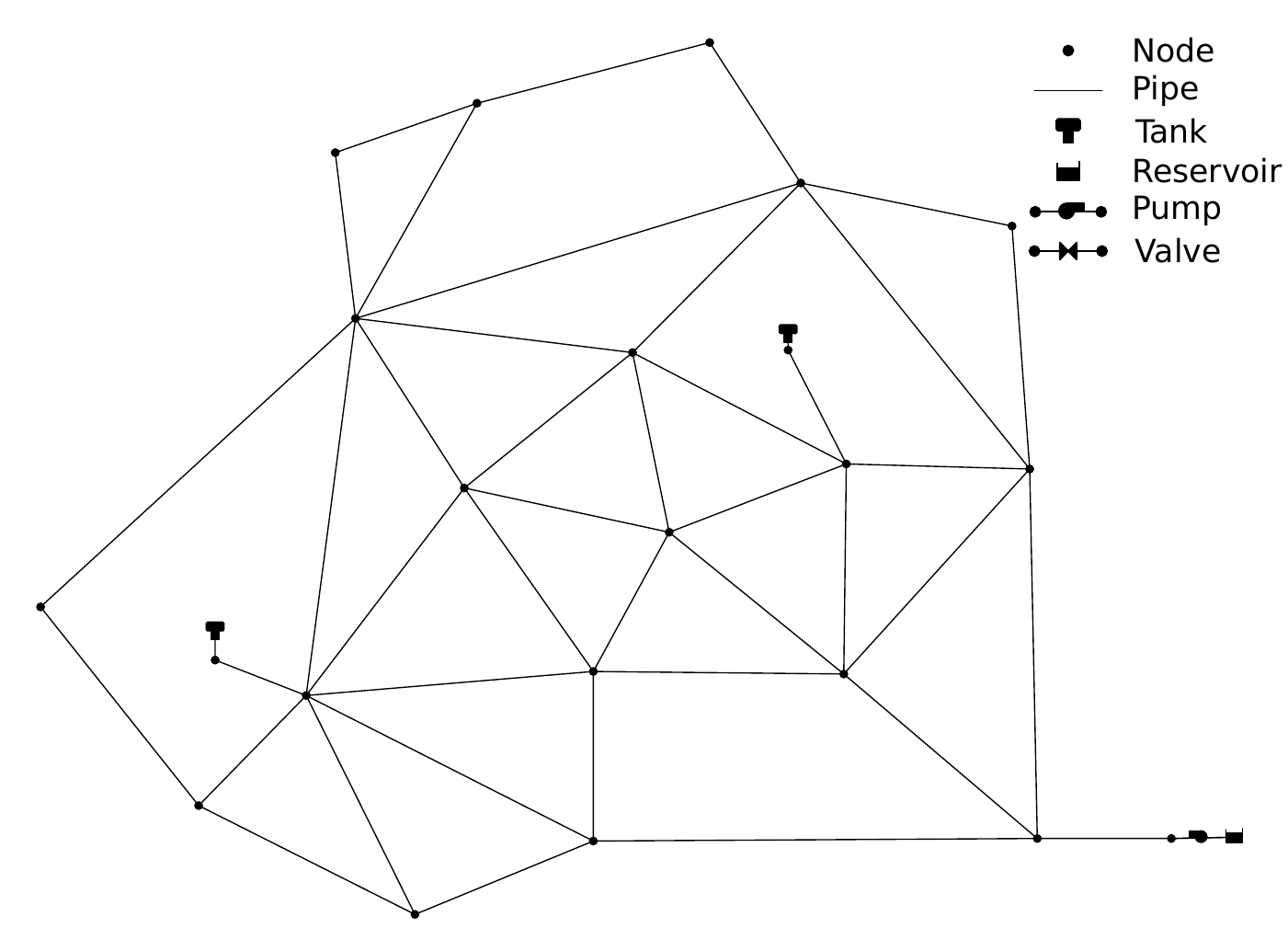}
    \caption{Topology of the Anytown water distribution system. Network elements plotted: nodes, pipes, pumps and tanks.}
    \label{fig:anytown_topo}
\end{figure}
\begin{figure}
    \centering
    \includegraphics[width=.9\linewidth, keepaspectratio]{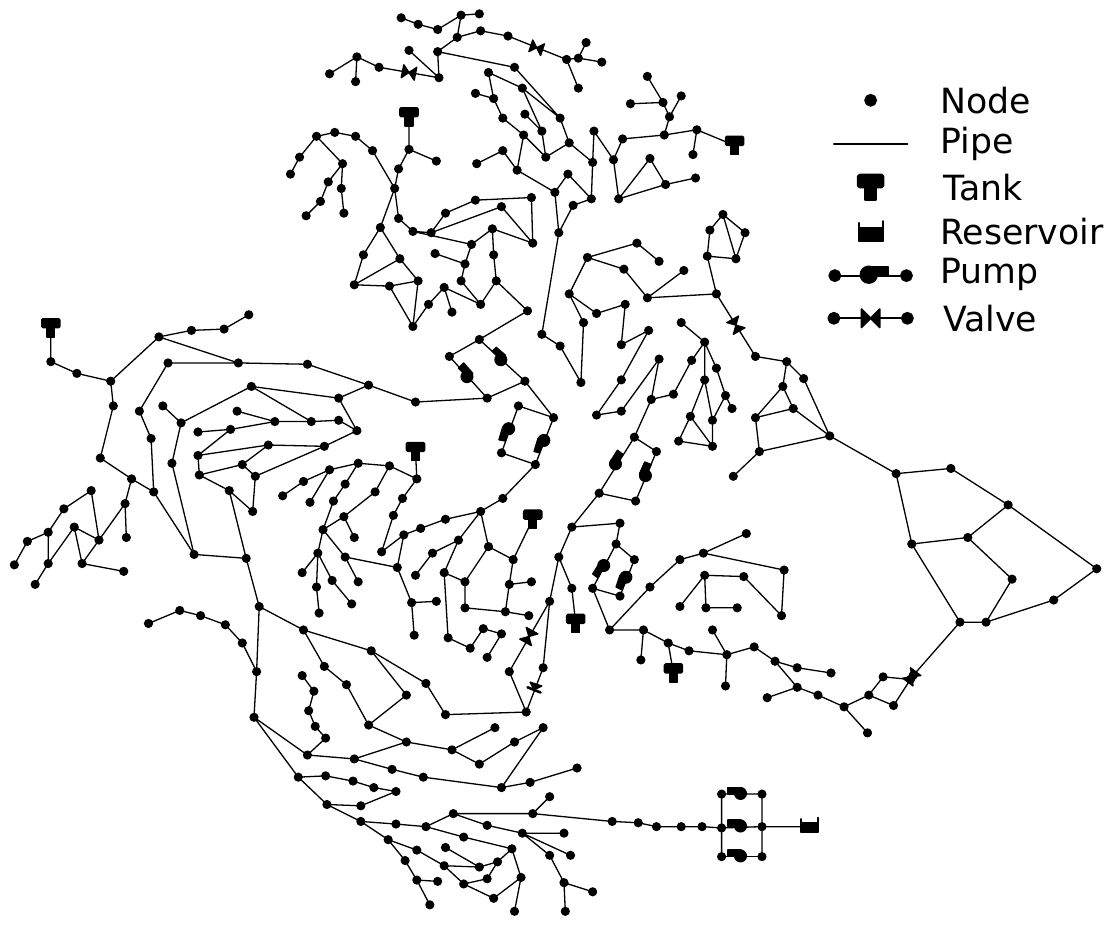}
    \caption{Topology of the D-Town water distribution system. Network elements plotted: nodes, pipes, pumps and tanks.}
    \label{fig:dtown_topo}
\end{figure}

The topology of the networks is depicted in Fig.~\ref{fig:anytown_topo} and in Fig.~\ref{fig:dtown_topo} for the Anytown and D-Town test cases, that are the same for Anytown-mod and D-Town-mod, respectively.
The networks are slightly modified as described below.
The primary purpose of Anytown-mod is to provide a simple, easily solvable problem that serves as a proof of concept: it has a global optimum that can be found with conventional methods easily.
If the agent performs poorly on Anytown-mod, then there is no chance that it can solve more complex tasks.
D-Town-mod is a more challenging problem whose solution is not trivial because of the multi-dimensional parameter space of the optimization loss function.
If the agent performs well on D-Town-mod, then the RL-based agent with the proposed reward function is a good candidate to serve as a real-time controller in real water networks.
The main properties of the two water distribution systems are listed in Table~\ref{table:benchmark_prop}.
Both water networks share their topology with its original version published by the \citet{cwsExeter}; the differences are summarized below.

\begin{table}
    \caption{Main properties of the water distribution systems}
    \label{table:benchmark_prop}
    \centering
    \small
    \renewcommand{\arraystretch}{1.25}
    \begin{tabular}{l l l}
    \hline\hline
    \multicolumn{1}{c}{Attribute of the water network} &
    \multicolumn{1}{c}{Anytown-mod} &
    \multicolumn{1}{c}{D-Town-mod} \\
    \hline
    Number of nodes & $22$ & $399$ \\
    Number of pipes & $41$ & $443$ \\
    Number of pump stations & $1$ & $5$\\
    \hline\hline
    \end{tabular}
    \normalsize
\end{table}

Anytown-mod is almost identical to Anytown as only the number of pumps in the single pumping station is reduced to two from three, and both of the pumps operate with the same continuously variable speed.
The performance curves of the pumps are the same as in the original network.

In the case of D-Town-mod, the control logic is removed, and all pumps operate in the water distribution network.
The performance curves are kept for the head, and fictitious curves are added for the efficiency.

An environment is created that serves as an interface between the agent and the water distribution system.
The purposes of the environment are the interpretation of the commands of the agent, reward calculation and re-initialization of the water network.
The state of the water network can be initialized by randomizing the demand map and by randomizing the pump speeds or resetting them to the original value.
The original demands in the task definition of the water network are stored, and the total sum of the original nodal consumption is calculated and stored at the beginning of training.

Demand variations are substituted with randomized demand distributions over the water distribution system.
This can be done because of training the agent with an off-policy method.
It is not necessary to show real-life demand variations to the agent in the training phase as it is trained off-policy.

The randomization of the nodal consumption is done in three steps to be able to mimic the total demand variation during day and night and the nodal demand variation as well.
First, new demand is generated in every node as the product of the original demand and a random value from a truncated normal distribution.
Second, new total demand is generated as the product of the original total demand and a random value from a uniform distribution.
Third, the nodal consumptions are rescaled such that the sum is equal to the one generated in the second step.
The process is summarized in Alg.~\ref{alg:consumption_init} where the set of original nodal consumptions and the randomized nodal consumptions are denoted by $\matr{C}_0$ and $\matr{C}$, respectively.

\begin{algorithm}
\caption{Initializing nodal consumptions}
\label{alg:consumption_init}
\begin{algorithmic}[1]
    \Function{InitNodalConsumption}{$C_{0}$}
        \State{$\matr{C} \gets \matr{C}_{0}$}
        \State{$c_{\mathrm{tot}} \gets \Call{RndUniform}{{\scriptscriptstyle\texttt{lo}=0.3, \texttt{hi}=1.1}} \cdot \sum{\matr{C}_{0}}$}
        \State{$c_{tot^*} \gets 0$}
        \ForAll{$c \in \matr{C}$}
            \State{$c \gets \Call{RndTruncNormal}{{\scriptscriptstyle\mathrm{mean}=1.0, \mathrm{stddev}=1.0, \texttt{lo}=0.7, \texttt{hi}=1.3}} \cdot c$}
            \State{$c_{tot^*} \gets c_{tot^*} + c$}
        \EndFor
        \ForAll{$c \in \matr{C}$}
            \State{$c \gets c \cdot \frac{c_{tot}}{c_{tot^*}}$}
        \EndFor
        \State{\Return{$\matr{C}$}}
    \EndFunction
\end{algorithmic}
\end{algorithm}

This randomization method allows control on the demand map both in relative and in absolute terms.
The nodal consumption randomized one-by-one mimics the relative variation of the demand map.
Meanwhile, the overall change in consumption through the day-night cycle is imitated by the variation of the total demand.
I.e., the total demand of the area covered by the water distribution network is considered first; then, the total value is partitioned among the nodes.

\begin{figure}
    \centering
    \includegraphics[width=.8\linewidth, keepaspectratio]{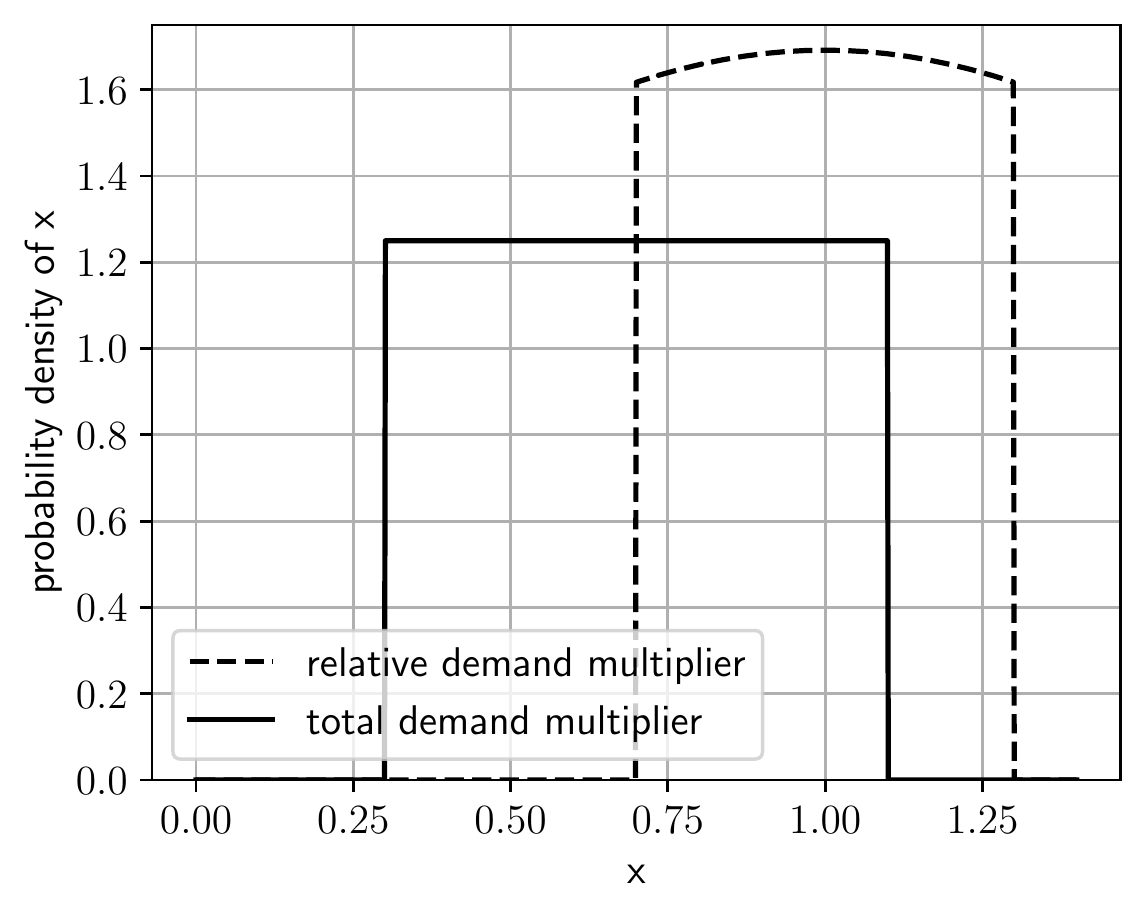}
    \caption{Probability distributions for the three-stage demand randomization.}
    \label{fig:proba_dists}
\end{figure}

Both distributions are chosen arbitrarily and their probability density functions are depicted in Fig.~\ref{fig:proba_dists}.
The lower and upper bounds of the uniform distribution are $0.3$ and $1.1$, respectively, while the normal distribution has a mean $1.0$ and standard deviation $1.0$ with tails cut off below $0.7$ and above $1.3$.
The shape of the truncated normal distribution is similar to the uniform distribution with this parameter set, although multipliers close to one have slightly higher probabilities during the demand randomization than remote values.

The pumps feeding the water network are considered as variable-speed pumps driven by e.g., frequency converters, in case of both water networks.
The absolute pump speed is not considered; the ratio compared to the nominal pump speed is recorded instead.
In the current environment, the speeds can be lowered to half of the maximum speed as water transportation is infeasible at lower speeds due to insufficient pump heads relative to the geodetic height of the nodes.
The cause of the lower speed limit is valid for both WDSs.

Finally, the well from where the pumps feed the network is considered as a reservoir with a water surface large enough to ensure a stationary water level during the speed-adjustment of the pumps.

\begin{figure}[ht]
    \centering
    \includegraphics[width=\linewidth,keepaspectratio]{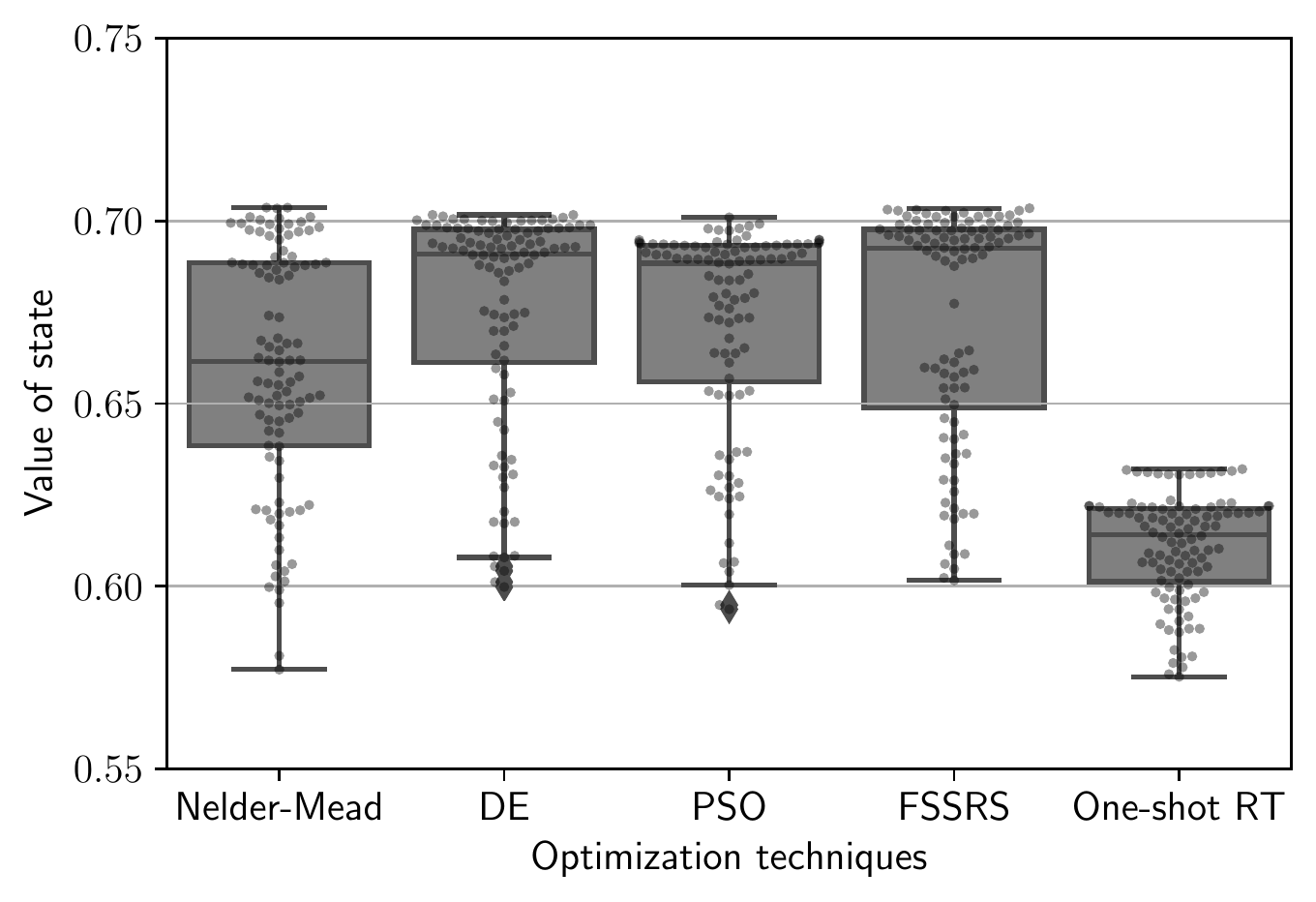}
    \caption{Results of the optimization on the test set with different techniques. The higher state values are the better.}
    \label{fig:opti_results}
\end{figure}

\begin{figure}
    \centering
    \includegraphics[width=\linewidth,keepaspectratio]{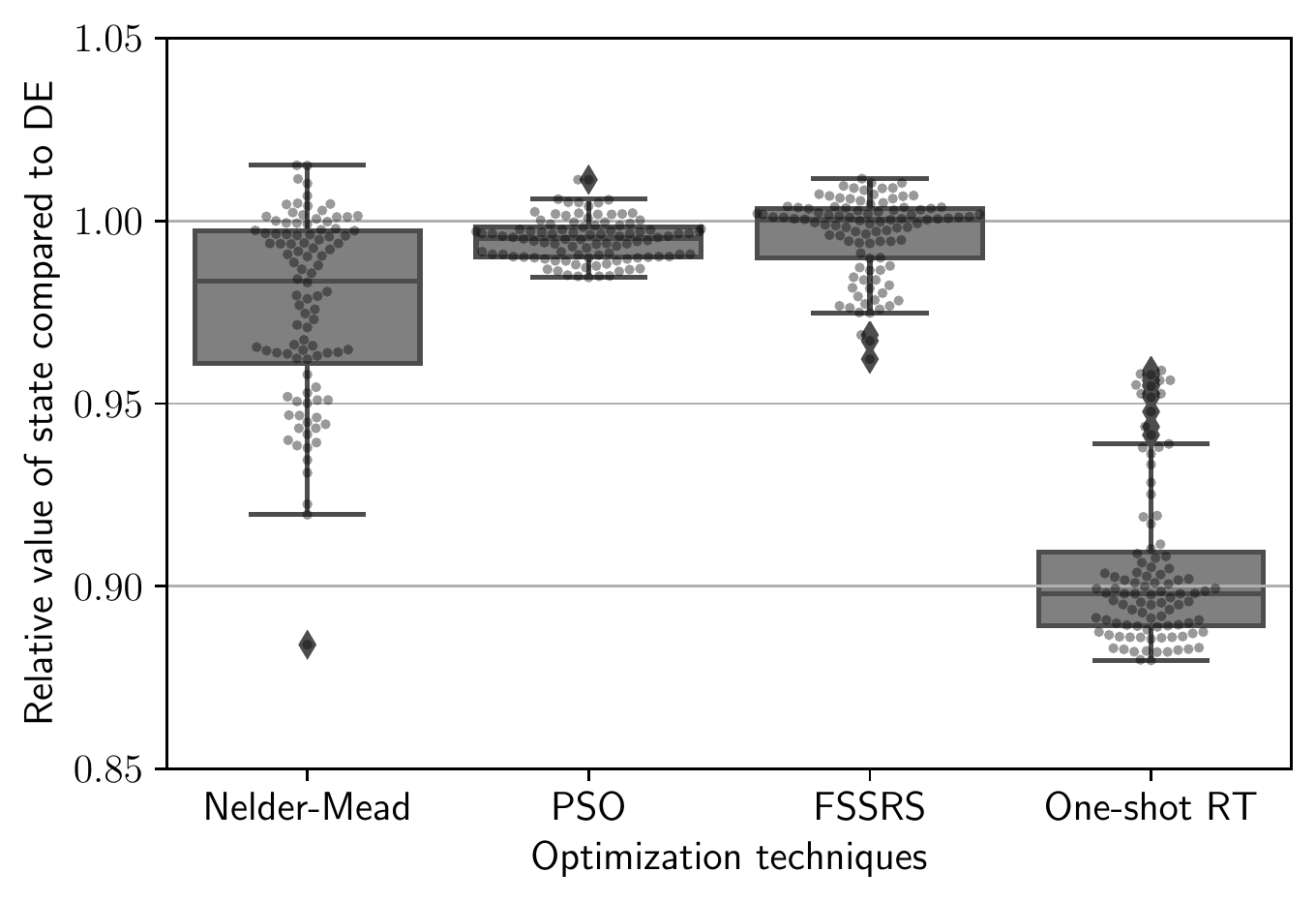}
    \caption{Box plots and swarm plots of the performance of the optimization algorithms compared to the differential evolution method. $1.00$ means the same performance that is achieved with DE.}
    \label{fig:opti_diffs}
\end{figure}

The presented results are obtained with the default hyperparameters of optimization algorithms.
In Fig.~\ref{fig:opti_results}, the average performance of the different algorithms is presented on the test set in the case of Anytown-mod and D-Town-mod.
Moreover, the box plot of the difference calculated for every test scenario is depicted in Fig.~\ref{fig:opti_diffs} where the results achieved by the differential evolution are chosen as a reference.

The similarity of the results suggests that each technique is able to find the empirical global optima, except for the one-shot random trial, which is reasonable.

The single pump station in Anytown-mod means that the pump speed optimization problem has a one-dimensional parameter space; hence, evolutionary algorithms and related techniques are inefficient in finding the solution as the crossover-like operations are meaningless on a one-dimensional candidate solutions.
Thus, only the performance achieved by the Nelder-Mead method was considered in the evaluation of scenarios from the Anytown-mod case.

\subsection{Training procedure of the agent}
\begin{table*}
    \caption{Possible hyperparameters for training}
    \label{table:ho}
    \centering
    \small
    \renewcommand{\arraystretch}{1.25}
    \begin{tabularx}{\textwidth}{>{\raggedright\arraybackslash\hsize=.25\hsize\linewidth=\hsize}X>{\centering\arraybackslash\hsize=.375\hsize\linewidth=\hsize}X>{\centering\arraybackslash\hsize=.375\hsize\linewidth=\hsize}X}
    \hline\hline
    \multicolumn{1}{c}{Hyperparameter} &
    \multicolumn{1}{c}{Anytown-mod} &
    \multicolumn{1}{c}{D-Town-mod} \\
    \hline
    Learning rate & \multicolumn{2}{c}{$\{10^{-4}, 10^{-3}, 10^{-2}\}$} \\
    Discount-factor & \multicolumn{2}{c}{$\{0.9, 0.99, 0.999, 0.9999\}$}  \\
    Batch size & $\{8, 16, 32, 64\}$ & $ \{16, 32, 64, 128\}$ \\
    \makecell[l]{\# of neurons \\ in the hidden layers} & \makecell{$\{[24]$, $[24, 6]$, \\ $[48,12]$, $[24, 12, 6]$, \\ $[48, 32, 12]\}$} & \makecell{$\{[300]$, $[300, 10]$, \\ $[128, 12]$, $[256, 32, 12]$, \\ $[256, 128, 12]\}$} \\
    \hline\hline
    \end{tabularx}
    \normalsize
\end{table*}

Two separate agents are trained during the experiment: one to solve the Anytown-mod problem and one to solve the D-Town-mod problem, respectively.

The agents were trained on the water networks separately, thus, during a training session, only one of the water distribution systems is utilized in the environment, and the trained agent will perform well only on the specific water network that was used for training.
The agents trained with different hyperparameters; the common part is just the underlying algorithm and the fact that a densely connected neural network is used as an action-value estimator.

The training will be described next.
At the beginning of every episode, the demands are randomized according to Alg.~\ref{alg:consumption_init}, the pump speed ratios are randomized according to a uniform distribution between the feasibility limits.
After every $\nicefrac{1}{25}th$ of the total number of training steps the training is suspended, and the performance of the agent is tested on $100$ different scenarios.
These scenarios are generated beforehand the training and they serve as validation data.
Additional $50$ test scenarios are generated, these scenarios are solved by the agent only after the training is completed.

A hyperparameter optimization was carried out prior to the training with a tree-structured parzen estimator (TPE) \cite{NIPS2011_4443}.
The critical hyperparameters were identified according to the previous experience of the authors.
Table~\ref{table:ho} summarizes possible hyperparameter values from where the optimal values were selected by the TPE.
Additional information on the sensitivity on the parameters and on the hyperparameter optimization is not presented due to length considerations.

The input layers of the neural networks have exactly as many neurons, as nodes in the water network as the nodal pressures divided by the highest shut-off head of the pumps are handed over the action-value estimator.
The activation functions are rectified linear units (ReLU) that are introduced in \citet{Nair2010}.
ReLU activation can handle input data outside the normalized range when the data are positive-valued, as is the case with nodal pressures under normal circumstances.
Nodal pressures are handed over to the neural network in vectors in every scenario, i.e., the $k^\mathrm{th}$ element of the pressure vector belongs to the same node under all circumstances.
The training is carried out with the $\varepsilon$-greedy policy starting with an exploration factor of $0.95$ and linearly decreased to $0.0$ at the final episode.
$\varepsilon$-greedy policy means that the agent makes random moves with a probability of the exploration factor, i.e., acting off-policy.
This behavior ensures the possibility to find better policies than the reference one, i.e., the agent is able to achieve high rewards even without mimicking the reference strategy.
The discount-factor affects the anticipatory behavior of the agent, the higher the value, the more prescient the agent is.
A scenario is $40$ and $200$ timesteps long in the case of Anytown-mod and D-Town-mod, respectively.
Moreover, the agents make moves respectively $50000$ and $1000000$ times in the smaller and the larger water distribution network before the final test.
The state-action-reward-new state data are saved to a so-called replay memory from where the training algorithm takes samples to fill the training batch for every update of the action-value estimator.
The hyperparameters of both agents are summarized in Table~\ref{table:hyperparams}.

\begin{table}
    \caption{Hyperparameters of the agents}
    \label{table:hyperparams}
    \centering
    \small
    \renewcommand{\arraystretch}{1.25}
    \begin{tabular}{l c c}
    \hline\hline
    \multicolumn{1}{c}{Hyperparameter} &
    \multicolumn{1}{c}{Anytown-mod} &
    \multicolumn{1}{c}{D-Town-mod} \\
    \hline
    Learning rate & \multicolumn{2}{c}{$0.0001$} \\
    Discount-factor & $0.99$ & $0.9$\\
    Batch size & $8$ & $64$ \\
    \# of neurons in hidden layers & $[48, 32, 12]$ & $[256, 128, 12]$ \\
    Size of replay memory & $25000$ & $350000$ \\
    Type of activation function & \multicolumn{2}{c}{rectified linear unit} \\
    Initial exploration factor & \multicolumn{2}{c}{$0.95$} \\
    Final exploration factor & \multicolumn{2}{c}{$0.00$} \\
    Maximum episode length & $40$ & $200$ \\
    Speed ratio increment & \multicolumn{2}{c}{$0.05$} \\
    \# of initialization steps & $1000$ & $10000$ \\
    \hline\hline
    \end{tabular}
    \normalsize
\end{table}

The hydraulic backend system is ensured by the numerical model of the water distribution systems, and these models are evaluated in \verb+EPANET+, which is a popular, open-source hydraulic simulator introduced by \citet{Rossmann1993}.
As the main parts of the computer code are written in \verb+Python+, only the core hydraulic calculations are computed by \verb+EPANET+, modification of the problem definition, and querying the results are conducted by the \verb+EPYNET+ interface that is a \verb+Python+ wrapper for \verb+EPANET+.
\verb+EPYNET+ is developed by IT experts at the Dutch drinking water company Vitens and is publicly available by \citet{epynet}.
The dueling deep q-network agent is a pre-packaged one from the reinforcement learning library \verb+Stable Baselines+ \cite{stableBL}.
The authors have developed the environment with which the agent interacts in \verb+Python+, and its interface is compatible with \verb+OpenAI gym+ for the seamless pairing of pre-packaged agents with the same standard interface.
Hyperparameter optimization was carried out with the \verb+Optuna+ package \cite{Akiba2019} for \verb+Python+.

\section{Evaluation and Results}
\begin{figure}
    \centering
    \includegraphics[width=\linewidth, keepaspectratio]{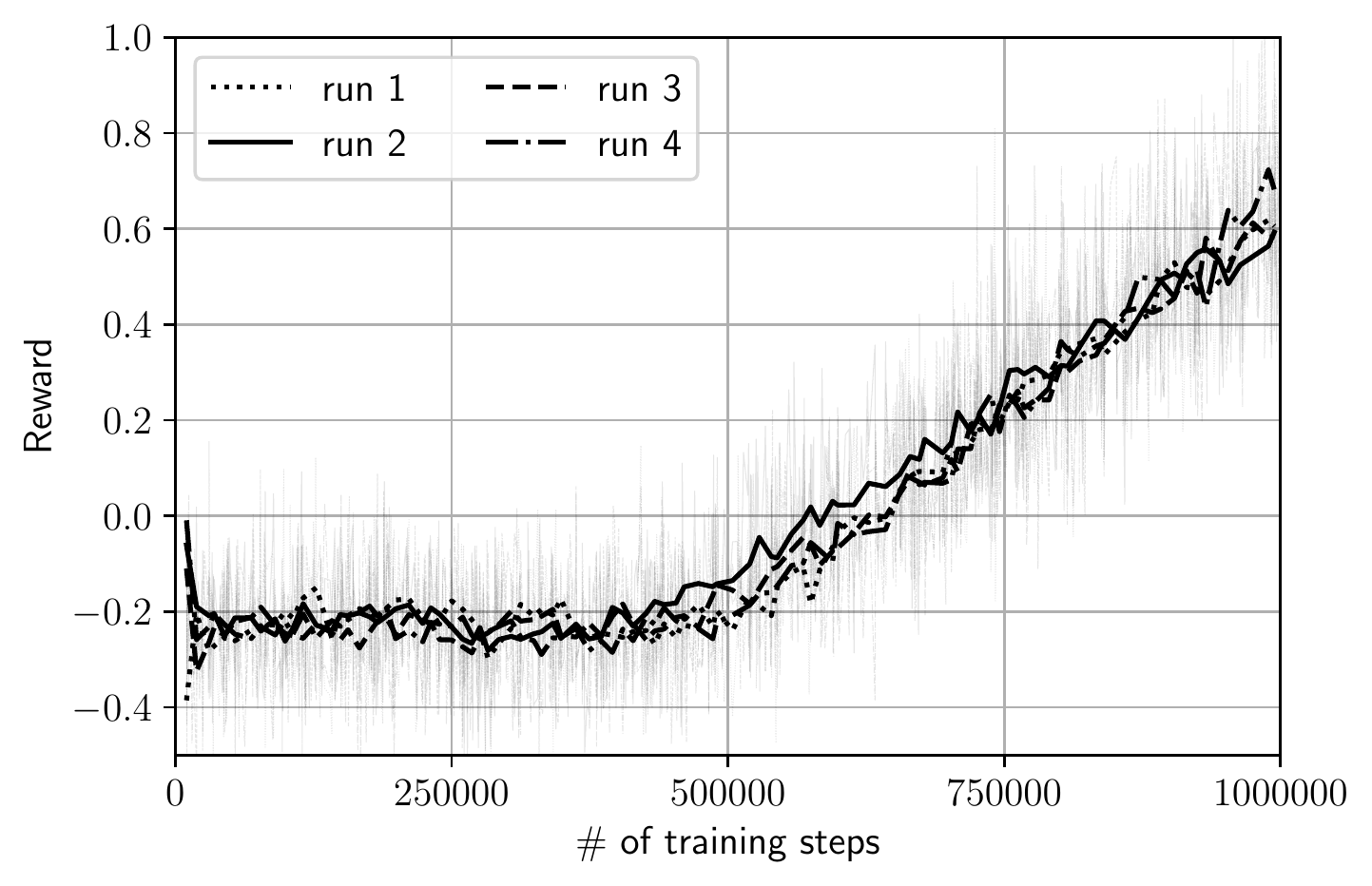}
    \caption{Reward during the training at D-Town-mod.}
    \label{fig:trn_rwd_dtown}
\end{figure}

\begin{figure}
    \centering
    \includegraphics[width=\linewidth, keepaspectratio]{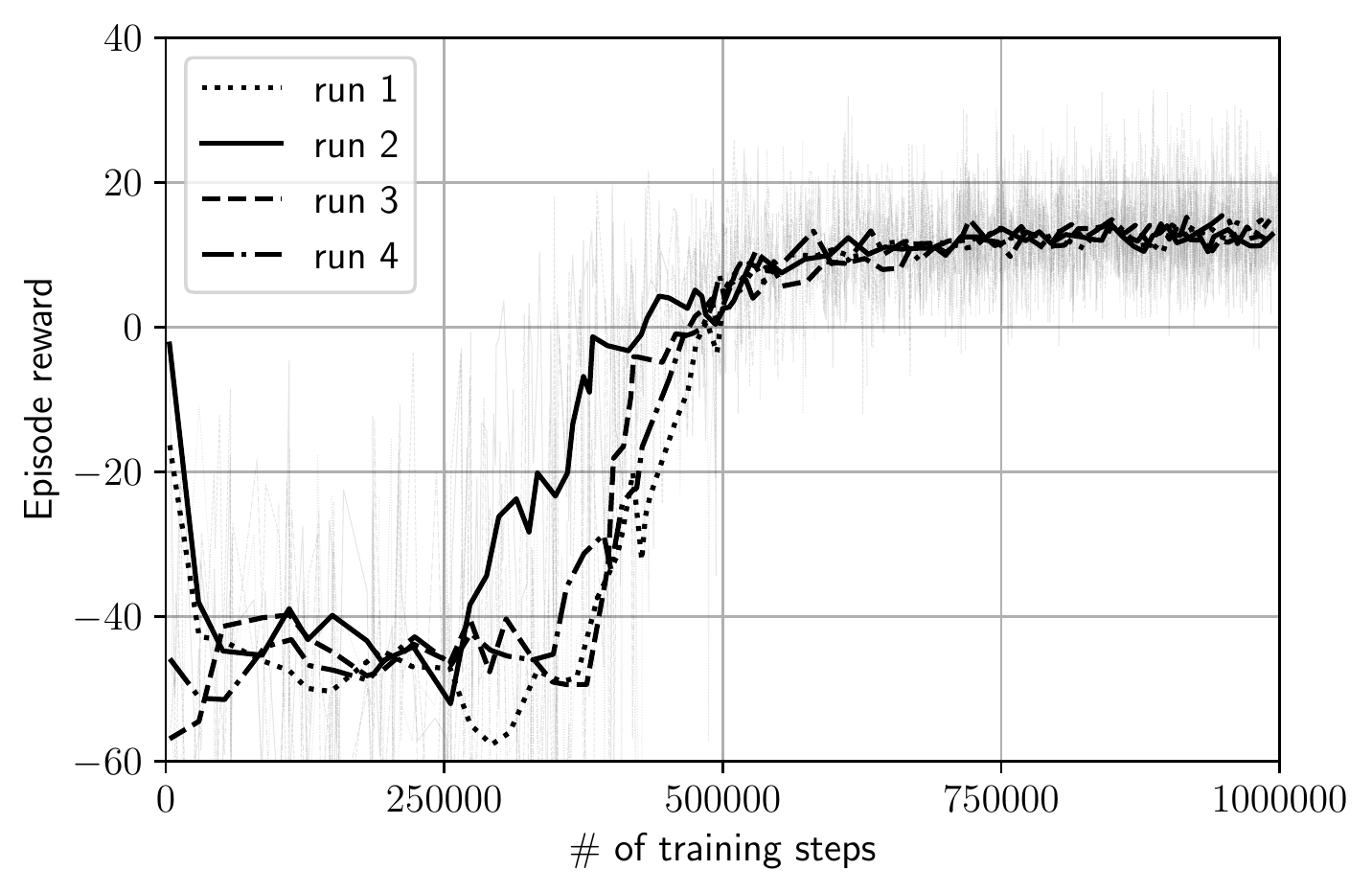}
    \caption{Episode reward during the training at D-Town-mod.}
    \label{fig:trn_ep_rwd_dtown}
\end{figure}

The history of the rewards and episode rewards is presented first for the D-Town-mod case in Fig.~\ref{fig:trn_rwd_dtown} and Fig.~\ref{fig:trn_ep_rwd_dtown}, respectively.
The history of the rewards and the episode rewards are similar in the Anytown-mod case; thus, those are not presented in the paper due to length considerations.
As the parameters of the action-value estimator are initialized randomly, the agent has no prior knowledge of the problem.
However, some initialization steps are taken by the agent to collect at least as many training samples with which the replay memory can be filled.
The underlying model -- the artificial neural network -- is updated first only after the initialization steps; thus, the training history is not available for the preceding steps.

Because of the random initialization, different training sessions can produce different results.
Moreover, the exploration capability of the agent introduces some stochastic behavior in the test performance of the agent through training.
Thus, $4$ training sessions are presented for each problem, and the results are smoothed by exponential moving averaging at $\alpha=0.3$.
As the agent starts taking steps according to a random policy, the achieved rewards are low at the beginning of the training.
The randomness of acting is maintained with a decreasing magnitude as the training goes on; thus, the rewards should increase as the action-value estimator is updated with more and more training samples.
Both the immediate and episode rewards saturate at the end of the training indicating that the action-value estimator (the deep q-network) cannot be developed further.
However, the ability of the agent to solve the task cannot be judged by only the training rewards because if the reward is improperly specified, the optimum-looking behavior can be related to a completely false policy from the viewpoint of the original objective.
Evolving false policies is a common problem in optimization tasks as most of them are exposed to the curse of Goodhart's law, see e.g. the work of \citet{Manheim2018} and \citet{Lehman2018}.

\begin{figure}
    \centering
    \includegraphics[width=\linewidth,keepaspectratio]{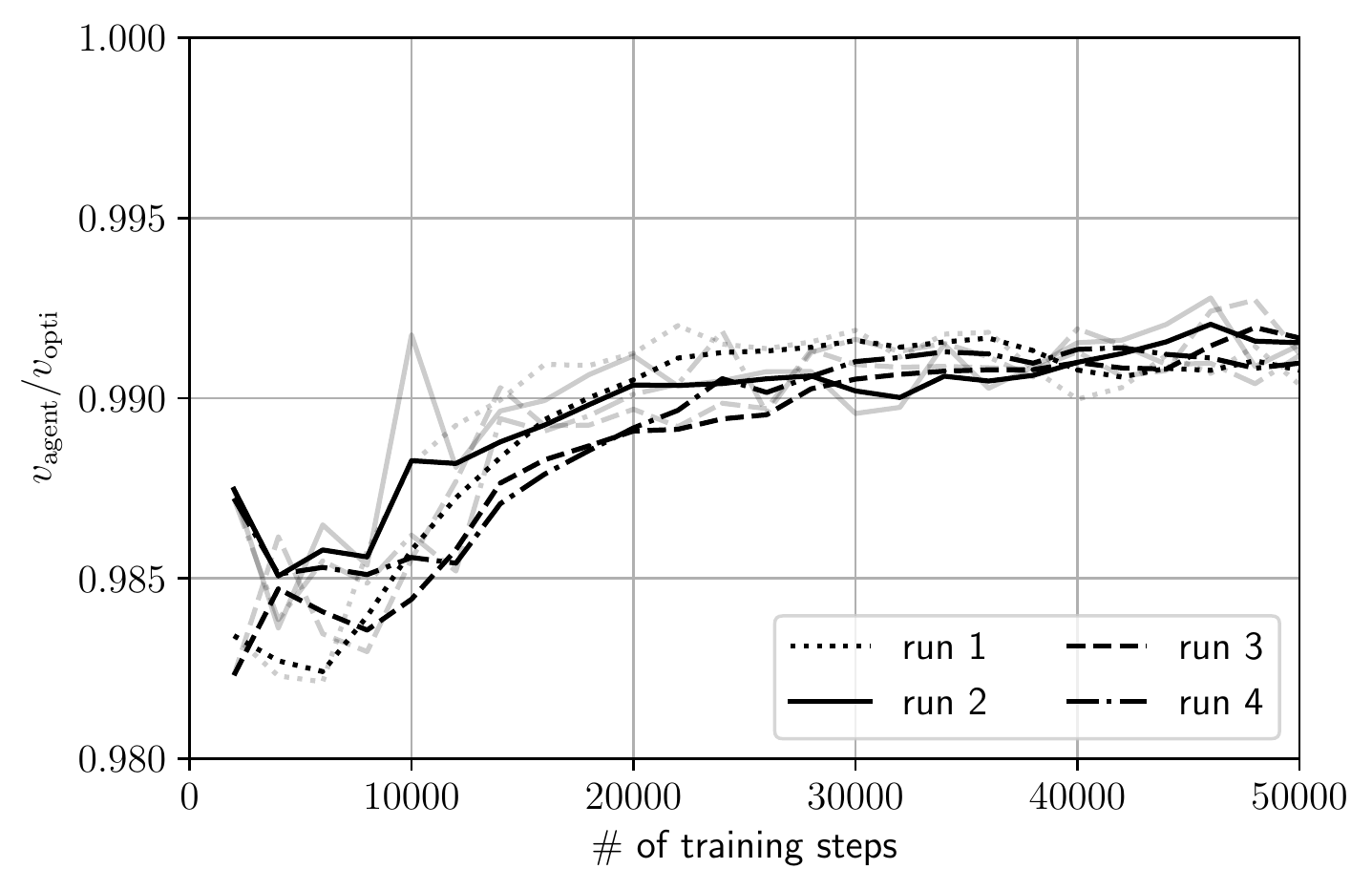}
    \caption{Average of the final state value achieved by the agent on the test set during the training at Anytown-mod.}
    \label{fig:val_rat_anytown}
\end{figure}

Thus, two additional metrics are examined in the agent's performance.
The most important aspect of the DDQN agent's behavior is whether it can reach comparable results to the reference solutions.
This property is examined on the test scenarios by looking at the terminal state value of the episodes denoted by the reference state value of the same scenario.
This is depicted in Fig.~\ref{fig:val_rat_anytown} for the Anytown-mod case.
The value ratio rises above $99\%$ in the first quarter of the training, and it stays there from then on in $3$ of the $4$ training sessions.
It implies that learning the optimal strategy in the sense of reaching high state value is relatively easy in the Anytown-mod case.

\begin{figure}[ht]
    \centering
    \includegraphics[width=\linewidth,keepaspectratio]{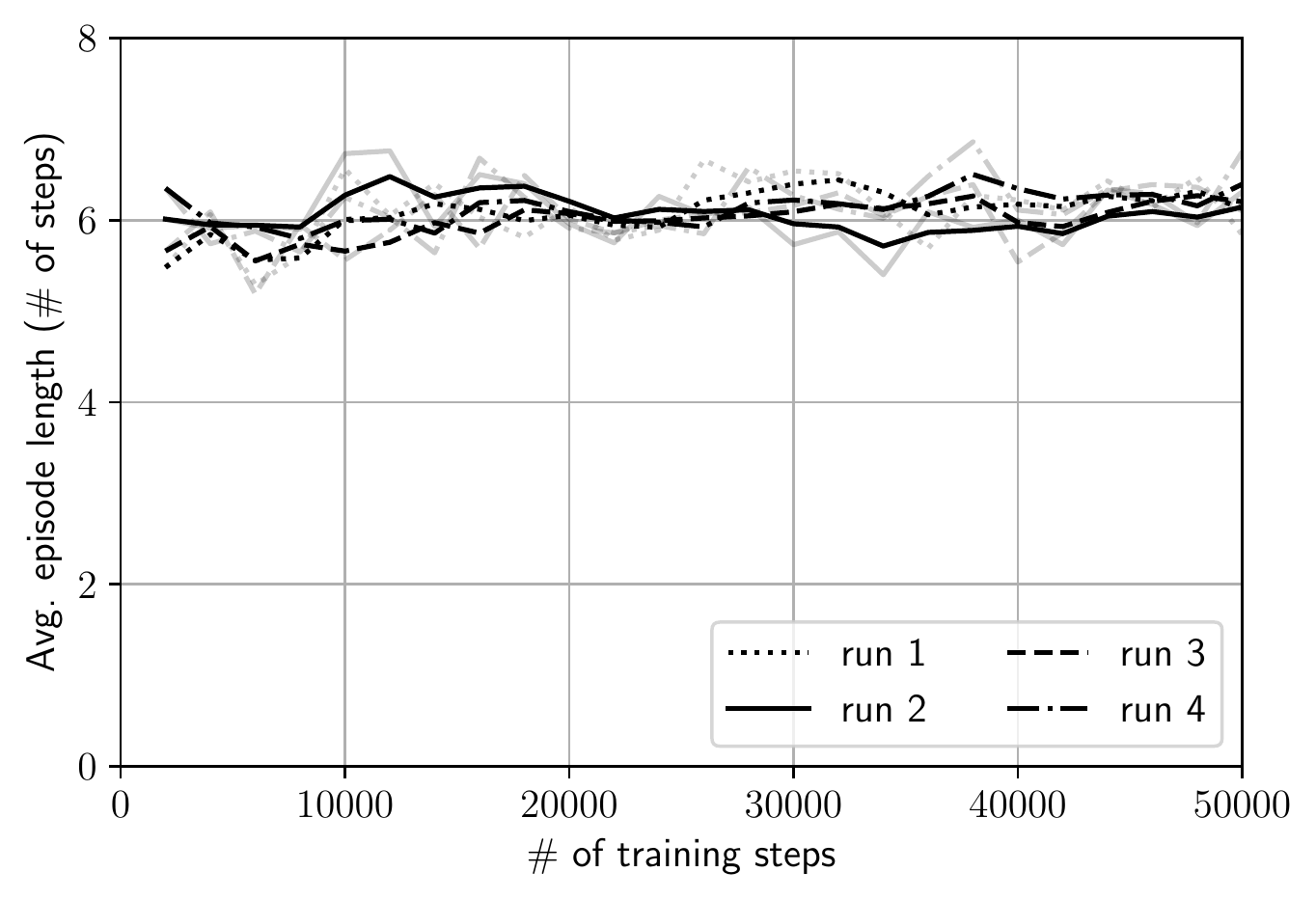}
    \caption{Average length of an episode in the test set during the training at Anytown-mod.}
    \label{fig:ep_len_anytown}
\end{figure}

However, the value ratio does not tell about the number of steps required to reach the terminal state.
Thus, the average length of the test scenarios are depicted in Fig.~\ref{fig:ep_len_anytown} for the Anytown-mod case.
The average number of steps taken before the termination is $6$, which means, that the agent takes at most $3$ active steps on average before terminating an episode with $3$ consecutive siestas.
As a reminder: there is only $1$ pump station in this environment whose speed ratio can be varied between $0.7$ and $1.1$ with increments of $0.05$.
Moreover, the initial speed ratio is randomized uniformly between the feasibility limits; thus a full sweep can be done in $8$ to $11$ steps with these settings (the initialization of the state is not counted as a step) and then the termination could be signaled with $3$ additional siesta steps.
Thus, $3$ active steps in the Anytown-mod means that the agent starts to move in the right direction.
The fact that both the high rewards and the shortest trajectory are achieved at the beginning of the training means that this amount of training steps are unnecessarily high for the Anytown-mod problem.

\begin{figure}
    \centering
    \includegraphics[width=\linewidth,keepaspectratio]{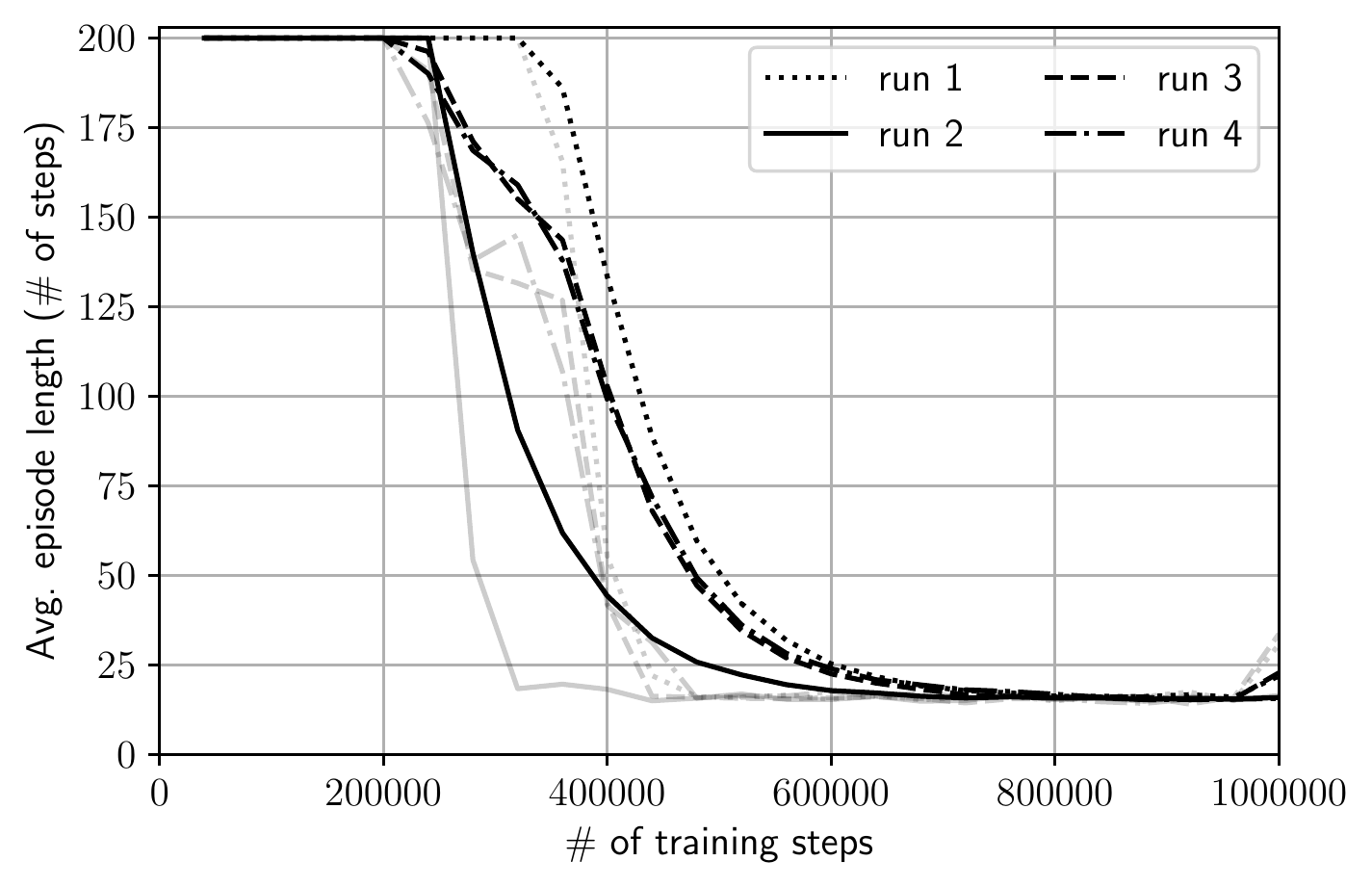}
    \caption{Average length of an episode in the test set during the training at D-Town-mod.}
    \label{fig:ep_len_dtown}
\end{figure}

\begin{figure}[ht]
    \centering
    \includegraphics[width=\linewidth,keepaspectratio]{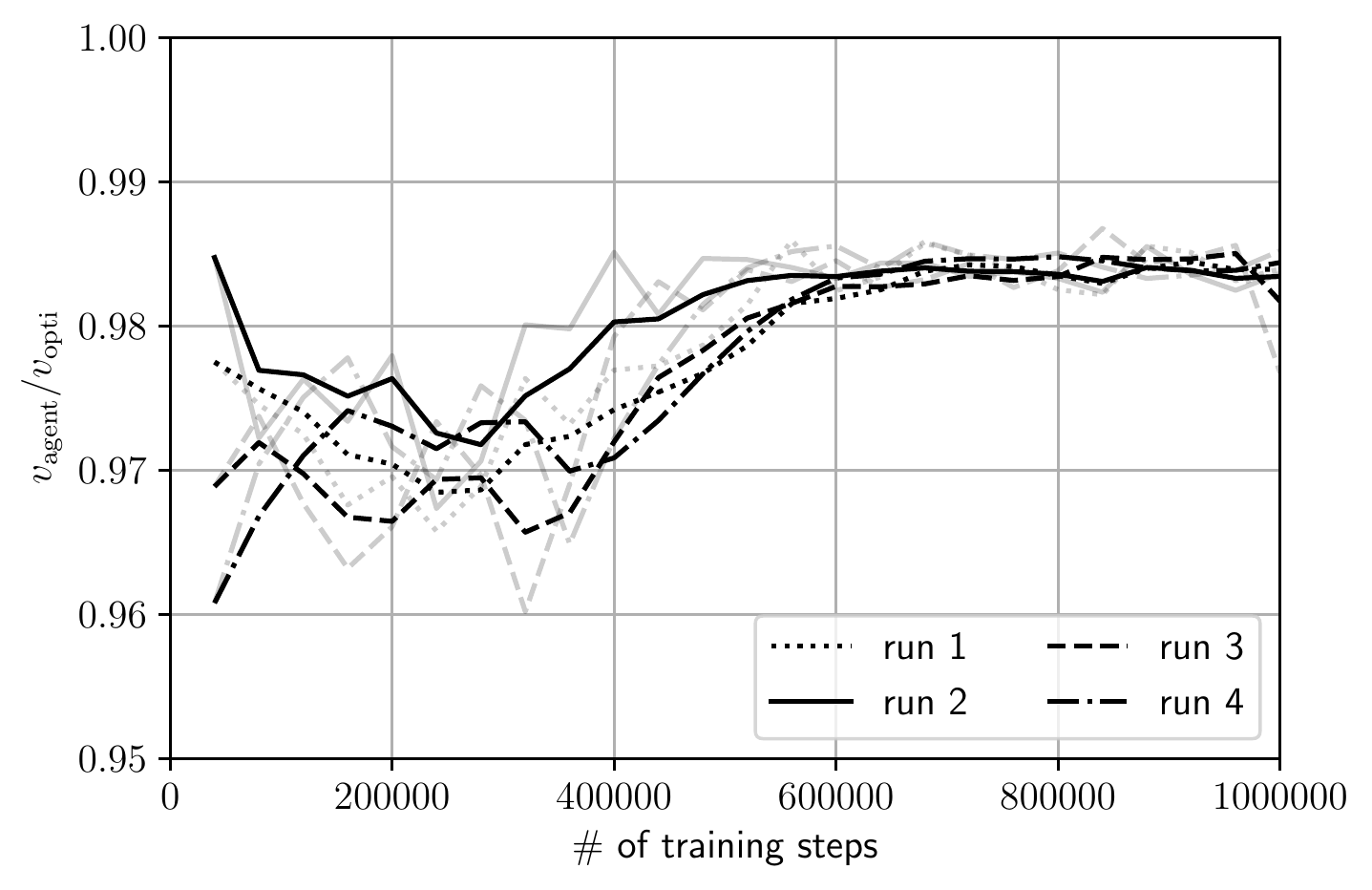}
    \caption{Average of the final state value achieved by the agent on the test set during the training at D-town-mod.}
    \label{fig:val_rat_dtown}
\end{figure}

The same metrics, i.e., average value ratio and average episode length over the test scenarios during the training are plotted from $4$ different runs in Fig.~\ref{fig:val_rat_dtown} and Fig.~\ref{fig:ep_len_dtown} achieved with the D-Town-mod environment, respectively.
According to Fig.~\ref{fig:val_rat_dtown}, DDQN also performs well in terms of value ratio in the more challenging environment.
However, the absolute performance is slightly lower than in the previous case: the agent achieves at most $98\%$ of the reference state value on average.
The average episode length differs from the previous case, where the shortest trajectories were found soon after the beginning of the training.
In this case, the agent exhausts all its possibilities to move around the parameter space until it learns that choosing a shorter trajectory leads to a higher reward.

\begin{figure}
    \centering
    \includegraphics[width=\linewidth,keepaspectratio]{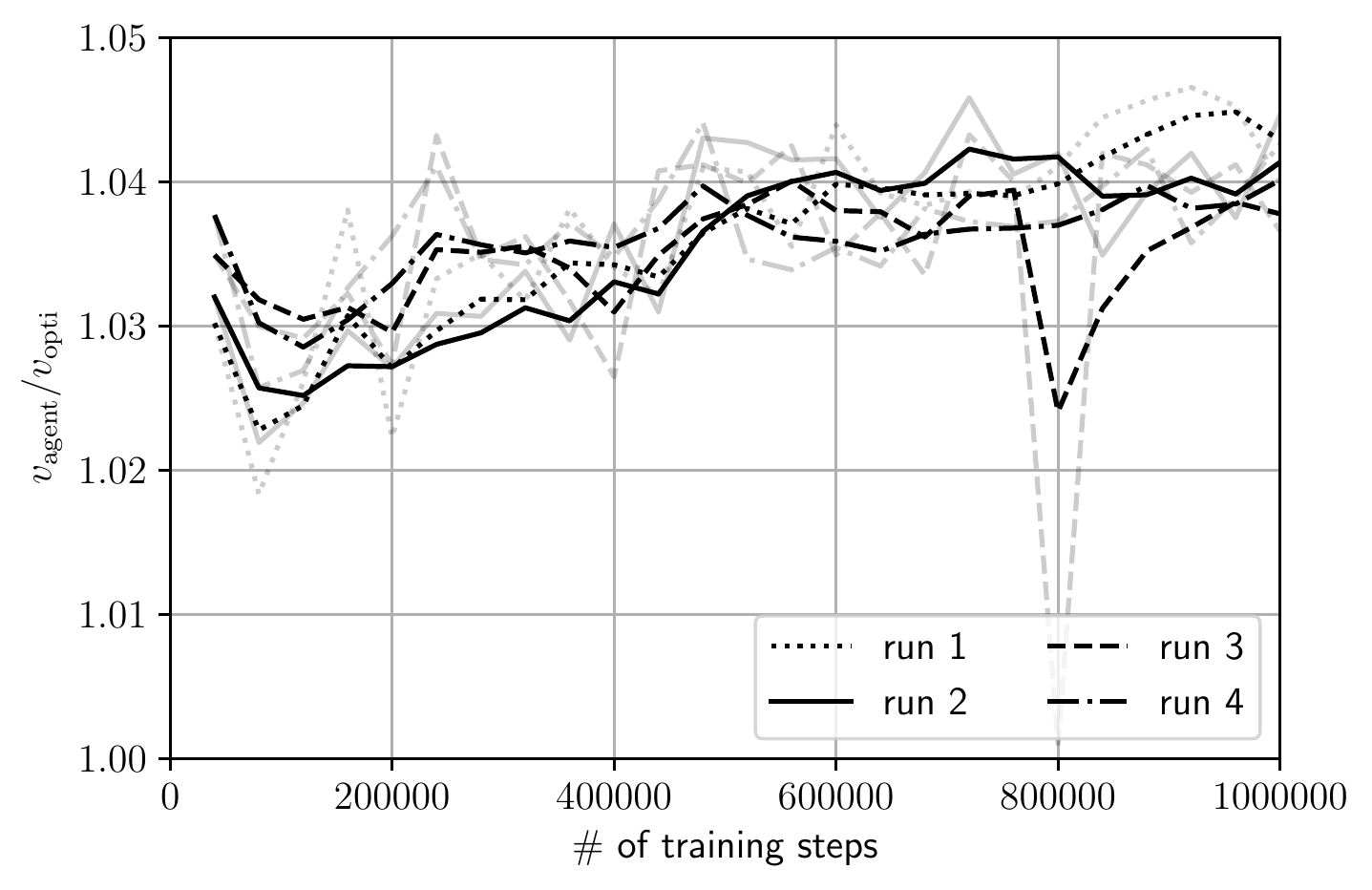}
    \caption{Average of the final state value achieved by the agent on the test set during the training at D-town-mod used one-shot random trial as a reference.}
    \label{fig:val_rat_dtownos}
\end{figure}

\begin{figure}
    \centering
    \includegraphics[width=\linewidth,keepaspectratio]{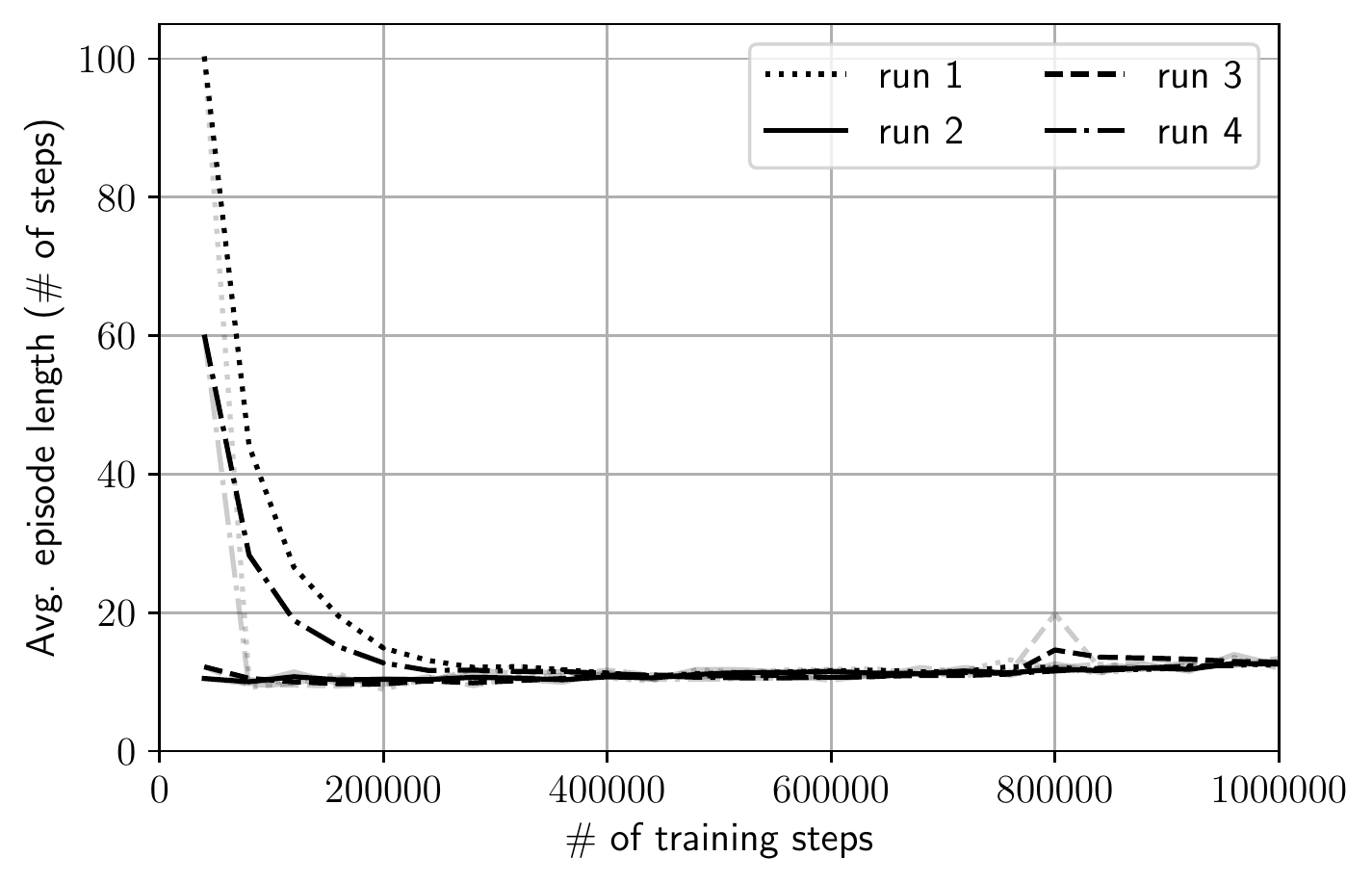}
    \caption{Average length of an episode in the test set during the training at D-Town-mod used one-shot random trial as a reference.}
    \label{fig:ep_len_dtownos}
\end{figure}

Finally, Fig.~\ref{fig:val_rat_dtownos} and Fig.~\ref{fig:ep_len_dtownos} represents the same performance metrics for the D-Town-mod environment but with the reference solution acquired with one-shot random trial.
So, in this case, the agent was guided by a suboptimal policy, but it can choose moves off-policy thanks to the fairly high exploration factor at the beginning of the training.
Owing to this, the DDQN agent performs reasonably well even with a poorly chosen reference algorithm as it outperforms that.


The average of the value ratios achieved on the test scenarios are summarized in Table~\ref{table:tst_results}.
The performance on the test set is similar to the the performance on the validation set in the end of the training that means the agent is able to generalize among different demand distributions.

\begin{table*}
    \caption{Agent performance on test scenarios}
    \label{table:tst_results}
    \centering
    \small
    \renewcommand{\arraystretch}{1.25}
    \begin{tabularx}{\textwidth}{>{\centering\arraybackslash\hsize=.10\hsize\linewidth=\hsize}X>{\centering\arraybackslash\hsize=.3\hsize\linewidth=\hsize}X>{\centering\arraybackslash\hsize=.3\hsize\linewidth=\hsize}X>{\centering\arraybackslash\hsize=.3\hsize\linewidth=\hsize}X}
    \hline\hline
    \multicolumn{1}{c}{run} &
    \multicolumn{1}{c}{Anytown-mod} &
    \multicolumn{1}{c}{D-Town-mod} &
    \multicolumn{1}{c}{D-Town-mod with one-shot RT} \\
    \hline
    1 & $0.991$ & $0.987$ & $1.04$ \\
    2 & $0.992$ & $0.987$ & $1.04$ \\
    3 & $0.991$ & $0.979$ & $1.04$ \\
    4 & $0.991$ & $0.988$ & $1.04$ \\
    \hline\hline
    \end{tabularx}
    \normalsize
\end{table*}

\section{Conclusions}
A methodology is presented for instantaneous pump control in water distribution systems with reinforcement learning.
The proposed technique is novel in the sense that it relies entirely on live measurement data in the decision-making process.
This property makes real-time optimal control of the pumps possible in a smart water network.

The underlying algorithm of the agent is a dueling deep q-network (DDQN).
As DDQN handles discrete action spaces only, the pump speed setting process is done by changing the speeds with fixed-size increments.

The performance of the agent is measured on two water networks with significantly different number of nodes, pipes and pump stations.
Metrics are defined as value ratio and episode length.
The former measures the value of the terminal state compared to the reference found by the Nelder-Mead method, while the latter shows whether the agent uses the shortest path to the optimum or is it just randomly seeking over the parameter space.
The novel approach is found to be useful, as the agent can achieve comparable results to widespread optimization algorithms even with the restrictions mentioned above.
In the more challenging case, the agent gets close to the reference solution with a difference in value ratio smaller than $2\%$.
As the agent gets an extra bounty if it reaches a value ratio higher than $98\%$, this performance ought to be upgraded by fine-tuning the rewarding function.
Another critical aspect of the agent-based pump control is whether the agent can do its task in a limited number of time steps.
The proposed rewarding algorithm is found to be appropriate to incite the agent to use the shortest possible trajectories towards the optimal state.
In other words, the agent was able to solve the more challenging task in around $20$ timesteps ($n_\mathrm{DQN}$) compared to the $70$ evaluations ($3.5 n_\mathrm{DQN}$) of the Nelder-Mead method and the few hundred evaluations ($\approx\!10 n_\mathrm{DQN}$) of the evolutionary methods.
Besides the number of evaluations, the classic methods rely on the hydraulic model of the water network, but the proposed technique is able to rely only on measurement data after training.

Moreover, the exploratory behavior during the training provides an opportunity to outperform the reference algorithm that is used for the training guide.
This phenomenon is also observed with the presented technique in the case when suboptimal guidance is used.

The proposed technique is a novel operation point optimizer for variable speed pumps in the presented state that depends only on measurement data during control.
It is also suitable for hybrid pump scheduling methods to assign optimal set-points to VSPs in real-time.
As the deep q-network can be changed to handle temporal processes, the presented technique can be further developed to a real-time pump scheduling method for water distribution networks.

\subsection{Data Availability Statement}
\begin{itemize}
\item Some or all data, models, or code generated or used during the study are available in a repository online in accordance with funder data retention policies (Reinforcement learning algorithms implementation \cite{stableBL}, Python API for EPANET \cite{epynet}, original numerical models of Anytown and D-Town \cite{cwsExeter}, the code repository of the presented research including everything to reproduce the results \cite{rl-wds}.)
\end{itemize}

\section{Acknowledgements}
 The research presented in this paper has been supported by the BME-Artificial Intelligence FIKP grant of Ministry of Human Resources (BME FIKP-MI/SC) and by the National Research, Development and Innovation Fund (TUDFO/51757/2019-ITM), Thematic Excellence Program. Bálint Gyires-Tóth is grateful for the financial support of the Doctoral Research Scholarship of Ministry of Human Resources (ÚNKP-19-4-BME-189) in the scope of New National Excellence Program and of János Bolyai Research Scholarship of the Hungarian Academy of Sciences. 




\bibliographystyle{elsarticle-num-names}
\bibliography{rl-wds.bib}

\end{document}